\documentclass{article}
\usepackage{spconf,amsmath,graphicx}
\usepackage{amsfonts}
\usepackage{booktabs}
\usepackage{siunitx}
\usepackage[export]{adjustbox}
\graphicspath{ {./Figures/} }
\usepackage{float}

\usepackage{xcolor}
\definecolor{scott_color}{rgb}{1, 0 ,0}
\title{Explaining Automatic Image Assessment}
\name{Author(s) Name(s)\thanks{Thanks to XYZ agency for funding.}}
\address{Author Affiliation(s)}
\twoauthors
 {Maxwell Lisaius\sthanks{In completion of a Masters of Science}}
	{Western Washington University\\
	Department of Computer Science}
 {Scott Wehrwein\sthanks{Advisor}}
	{Western Washington University\\
	Department of Computer Science}

\begin{document}
%\ninept
%
\maketitle
\begin{abstract}
  Previous work in aesthetic categorization and explainability utilizes manual labeling and classification to explain aesthetic scores. These methods require a complex labeling process and are limited in size. Our proposed approach attempts to explain aesthetic assessment models through visualizing dataset trends and automatic categorization of visual aesthetic features through training neural networks on different versions of the same dataset. By evaluating the models adapted to each specific modality using existing and novel metrics, we can capture and visualize aesthetic features and trends.
\end{abstract}

\section{Introduction}
\label{sec:intro}

Assessment of image aesthetics is a subjective and difficult problem in computer vision. Recently, some success has been demonstrated using neural networks to evaluate the aesthetic quality of an image; however, these approaches usually lack the ability to explain what makes an image aesthetically good or bad. More explainable aesthetic assessment techniques could be useful in a variety of applications, including photography, both pre- and post-capture, social media analysis, discriminators in generative models, or for applications in the arts.
\\
In this paper, we examine a recent neural network-based approach to aesthetics assessment \cite{NIMA} and impose some level of explainability by training models on variously modified versions of the input image. We call these predefined image manipulations and decomposition techniques ``modalities''; their purpose is to extract meaningful subsets of visual information about the source image: for example, geometric composition via depth maps, and visual interest via saliency maps.
\\
Most prior methods attempting to provide explainable aesthetic ratings have relied on hand-designed labels or categories, often based on established ideas about aesthetics. For example, CADB \cite{CADB} and EVA \cite{EVA} which add labels for semantic and aesthetic information.

While our choices of modalities are hand-designed, our goal is to leave the mapping between image characteristics and aesthetic quality for the model to understand. For example, rather than assume that images following the Rule of Thirds are good, we would expect a model trained on the depth map modality to learn that such positioning of objects tends to result in higher aestheic scores.

To progress towards explainable aesthetic assessment, we reimplemented the Neural Image Assessment \cite{NIMA} method, explored several small modifications that improved its performance on consumer-grade hardware. Our first contribution comes in the form of metric re-evaluation, helping establish a baseline for new metric choices in aesthetic assessment. We then trained multiple versions of the model on three modalities: depth (an estimated depth map), saliency (an estimated saliency map), and blur (a lower-resolution, blurrier copy of the original). We compared the relative accuracy of these models on each modality and we were able to quantify trends and categorize the dataset using the modalities.

\section{RELATED WORK}

Image aesthetic and composition evaluation is an open research subject with ongoing applications and a history of interest in the field. Older methods that use saliency and rule of thirds based techniques \cite{oldphotocomp} used limited datasets and were implemented using algorithms. Other methods in related areas aim to recompose or improve images \cite{oldqualityassess}. While these methods have been shown to work, they have limited generalization and often make assumptions about inital photo composition. Later the AVA dataset \cite{AVA} was collected and labeled: it provided a large dataset to the task and is now the most common benchmark for the photo aesthetic analysis task.

With the introduction of a large dataset, it opened the gateway for use of neural networks.  Methods and papers like  DMA-Net \cite{DMA-Net} utilize convolutional neural networks to estimate aesthetic rankings of the AVA dataset. Further work lead to papers like NIMA \cite{NIMA} where improvements to the loss functions were made to better represent diversity in aesthetic rankings, as well as increasing accuracy and correlation scores. 

State of the art methods like MP\_adam \cite{MPADAM} and A-Lamp \cite{A-LAMP} make use of patches, attention mechanisms, and significant hardware to focus in on specific aesthetic features attacking accuracy metrics. Pool-3FC \cite{POOL3FC}  pools features in a comparable way to also squeeze the most amount of information out of the source data.

\section{METHODS}

\subsection{NIMA}

In 2018 NIMA \cite{NIMA} was published and introduced a new loss function to the task, earth movers distance, a loss function originally proposed in \cite{OGEMD}. Given that the dataset being used contained a distribution of scores it made sense to have the network also predict a distribution, therefore EMD was a logical choice to measure the distance between two distributions. NIMA applied this loss function to several architectures including Inception-v2 \cite{InceptionV2}, MobileNet \cite{MOBILENET}, and VGG16 \cite{VGG16}. Once these networks were trained they were able to apply their methods to aesthetic assessment and image enhancement tasks.

\subsection{Re-implementation and modifications}

We started with the loss function and architecture used in NIMA and made a series of small modifications to improve the results in meaningful ways with the hardware available to us. We chose NIMA as a good starting point because it shows success in capturing aesthetic relationships without imposing attention methods like in A-Lamp \cite{A-LAMP} or \cite{MPADAM}. While these methods improve on accuracy scores, we did not want to introduce attention methods as an extra variable and opted to select a well established understood architecture. As mentioned previously they applied their methods to multiple CNN backbones, this allowed us to select an architecture that was a good fit for the consumer-grade hardware available to us.

By applying this same model to our decomposed modalities and measuring each models' accuracies in multiple metrics, we can discern meaningful underlying information about the image content and the dataset itself. While generating new datasets and training multiple models is not as straightforward as providing more descriptive labels, the process can be done automatically and is free from biases imposed from the categorization re-labeling process.

Initially we tried different CNN backbones with EMD loss. In these observations, the VGG\cite{VGG16} network was the most welcoming to train, achieving better results compared to other modern architectures when applied to the task. The VGG network is also relatively smaller in size and simple to apply weight freezing to, making it a good network to create a baseline to start with. 

One of the underlying challenges of aesthetics datasets as mentioned in previous work is the tendency for the predicted scores to be clustered around the mean. When training neural networks with a dataset that contains this property, the resulting model will also predict the same, both in individual predictions, as well as the overall distribution of scores. When a model trained like this is given data it is indecisive on, the default prediction will match closely to that of the overall dataset. The result of this activity is that to achieve the best accuracy and loss scores the model will have a high clustering of scores around the dataset mean.

In our training we made modifications to the EMD loss function that would encourage the model to weigh datapoints further away from the central cluster more heavily. Our loss function adds two additional terms multipled onto the EMD loss that can be tuned as hyperparameters. These terms take into account the mean and variance of the predicted aesthetic distribution comparing it to the ground truth for that data point. Allowing these hyperparameters to be tuned provides the model with the ability to weigh each datapoint based on its distance from the dataset mean, in addition to the error in the prediction.

Similar loss terms have been used in the past for age estimation in faces \cite{meanvarloss}. However for their application the difference between the ground truth and predicted mean and variance was used, and the method was applied for different effect, to "pull in" or "concentrate" the distribution of predicted ages around the ground truth. In the paper \cite{meanvarloss} they show that the technique also translates well to aesthetic assessment. The main difference between their loss and the terms we added to EMD is they compare their prediction to the ground truth as is common in loss functions, while ours compares the prediction to the dataset variance and mean.

Our combined loss function is as follows:
\\
\[ \text{Loss} = \text{EMD}(p, \hat{p}) * h_{\mu}  \Delta_\mu * h_{v}  \Delta_{\sigma ^2} \]
\\
Where $h_{\mu}$ and $h_{v}$ are hyperparameters. 
\\
\\

 The EMD term is defined as: 

\[\text{EMD}(p, \hat{p}) = \left( \frac{1}{N}  \sum_{k=1}^{N} | \text{CDF}_p (k) - \text{CDF}_{\hat{p}}(k) |^r \right)^{1/r} \]
\\
Where $\text{CDF}_{p}(k)$ is the cumulative distribution function as $\sum_{i=1}^{k}\bf{p}_{s_{i}}$. Our loss terms $\Delta_{\mu}$ and $\Delta_{\sigma^2}$ are defined as follows: 
\\
\[ \Delta_\mu = 1 +  | 5.3833 - \mu(p) |  \]
\\
\[ \Delta_{\sigma ^2} =  1 + | 2.084 - \sigma ^2(p) |  \]
\\

Where the constants 5.3833 and 2.084 are the average mean and variance of the ground-truth distributions in the dataset. As in previous work \cite{NIMA}\cite{OGEMD} $r = 2$ allowing for easier optimization with gradient descent. For all validation and metric purposes, the original EMD loss function was preserved so that hyperparameters have no effect on reported EMD metrics. 

\subsection{Modalities}

In addition to applying the NIMA reimplementation to the original RGB images in the dataset, we explored its performance on different modifications of each image, called ``modalities''. We selected several image modifications that capture various different aspects of the underlying image content: Saliency, Depth, and Blur. 

Visual saliency is a natural choice, having been used in prior aesthetic assessment methods \cite{A-LAMP} to capture and provide focus to aesthetic content and detail in images. We used OpenCV's implementation of visual saliency to compute a saliency map for each original image in the dataset. Depth information captures geometric and spatial aspects of a scene without any texture information. Since most RGB images do not have ground truth depth maps, we created estimated depth maps using a single image depth estimation method \cite{MPI}. Finally, some images' aesthetics might be dominated by color choices, rather than fine details. For this we used a simple series of down sampling and bilateral filters, to produce a blurred image modality. The blur process starts with a 9x9 Gaussian blur kernel, followed by a bilateral filter. The bilateral filter uses a pixel neighborhood of 20 pixels, with a sigma value for the coordinate space and color space of 50. After that we then resize the image down to 32x32, apply another Gaussian blur with a 3x3 kernel, then resize the image back to its original size.

\begin{figure}[H]
\centerline{\includegraphics[scale=.22]{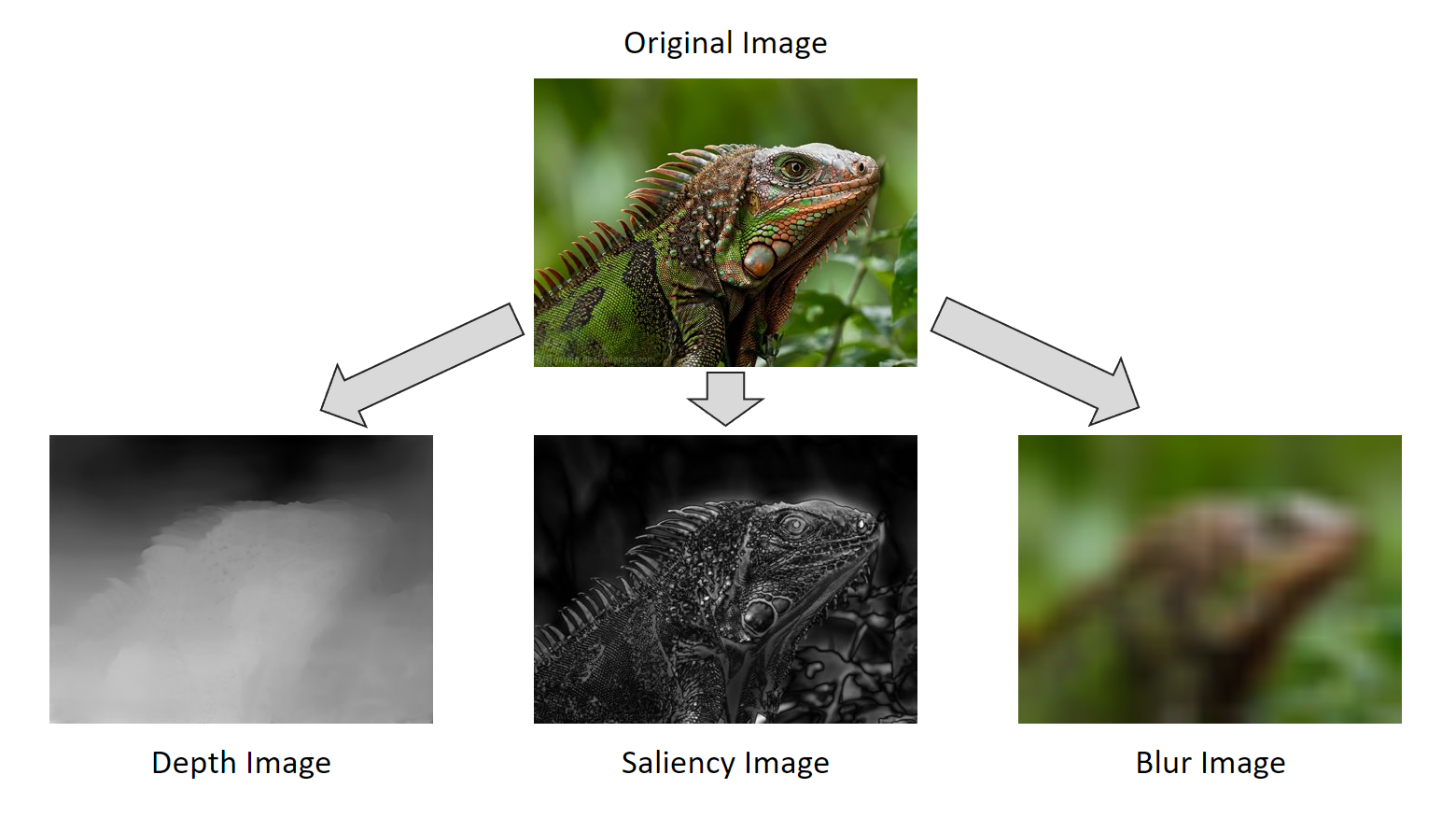}}
\caption{A breakdown of a source image into different modalities}
\label{fig:modsplit}
\end{figure}

\subsection{Transfer Learning}

Once the modality datasets were produced, we trained an image assessment network on each, to create a trained model specific to each modality. When training the different modalities, one of the biggest challenges was getting the modalities to begin a meaningful gradient descent with SDG. Often runs would get trapped in local minima and make no further progress. Due to the time and compute cost of these runs it became evident that a form of transfer learning from the RGB method would be helpful to start things in the right direction.

For this we implemented a method of freezing the network that would freeze a variable amount of the network layers. The freezing would start at convolution layers furthest into the network, and continued to freeze layers all the way until the input layer of the network. This method proved to be successful in allowing the early layers in the network to adjust from the RGB pretraining to the specific modality content. We varied the amount of frozen layers as a hyperparameter when tuning our models.

\begin{figure}[htbp]
\centerline{\includegraphics[scale=.8]{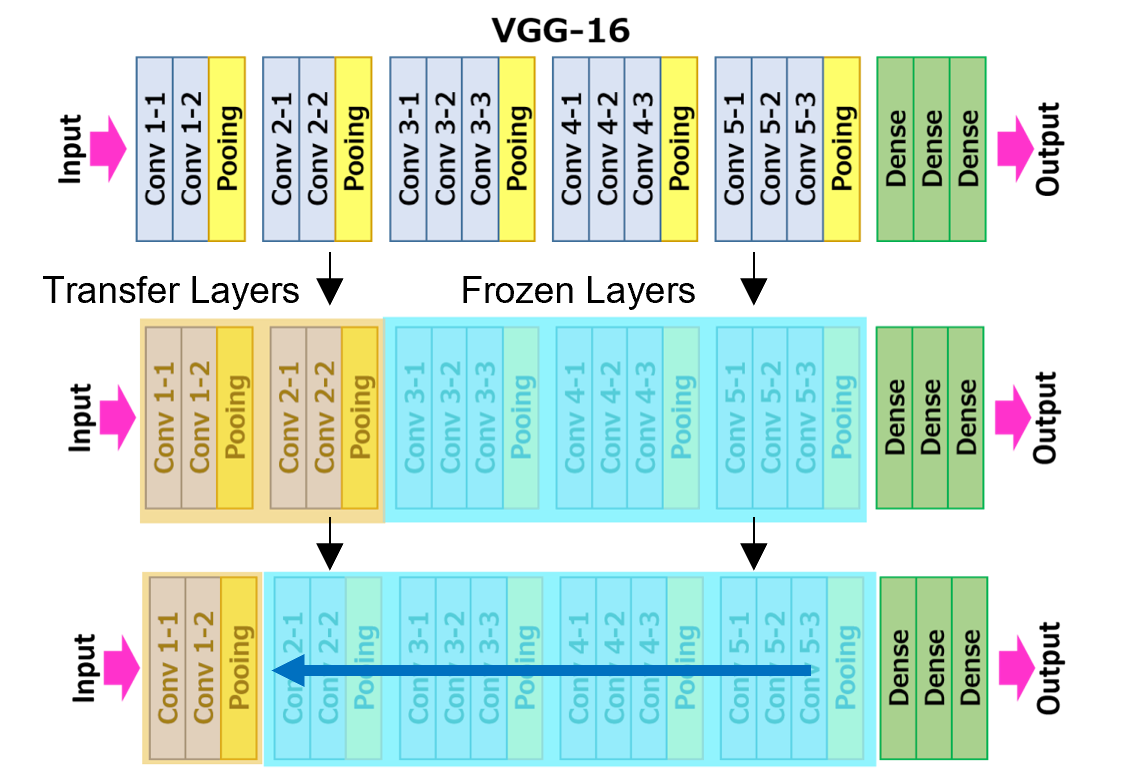}}
\caption{An illustration of the variable layer freezing methods used for transfer learning for early modality training.}
\label{fig:freeze}
\end{figure}

\section{EXPERIMENTS}

\subsection{Datasets}

The dataset we chose to apply our methods to is the most widely used image aesthetics dataset; AVA or Aesthetic Visual Analysis \cite{AVA}. This dataset contains over 255,000 images with many aesthetic assessment scores per image making it the largest dataset available for this task. These assessment scores are compiled into a 10-bin discrete distribution per image.

We evaluate and compare our model trained on the AVA dataset and its derived modalities to the EVA \cite{EVA} and AADB \cite{AADB} datasets. The EVA dataset is a subset of the AVA dataset that has been relabeled along with additional category labels to attempt to explain the aesthetic content. These categories include "color and light", "composition and depth", "quality", and "semantic" for aesthetic categories and for semantic categories they use "animals", "scenes", "human", "natural",  "still life", and "other". AADB is comprised of new images not contained in AVA, and additionally has category labels and metadata similar to EVA. AADB's categories include "balancing element", "color harmony", "content", "depth of field", "light", "motion blur", "object", "repetition", "rule of thirds", "symmetry", and "vivid color". A recent addition to the aesthetic dataset space is the CADB dataset \cite{CADB} that aims to categorize and evaluate the composition of a set of aesthetic images. We use these datasets with category labels to help understand the relative performance of our different modalities.

\subsection{Metrics}

The metric most commonly used for evaluating aesthetics models is the two-class percent accuracy, representing the binary decision of whether the image is ``good'' or ``bad''. The threshold between good and bad is usually chosen to be 5, the middle of the possible score distribution. Numbers like 80-83 percent are cited as state of the art \cite{A-LAMP} \cite{MPADAM}. A baseline accuracy with this threshold, however, is significantly higher than 50\% because the score distribution in AVA is not centered around 5; rather, the mean score in the AVA dataset is 5.38331. The difference of 0.38331 at first glance appears to be an insignificant amount, however due to the strong clustering of the model's predictions around the dataset mean, when setting a threshold, this amount becomes far from insignificant: as seen in Table \ref{fig:2class}, a baseline that predicts that all images are ``good" achieves over 70\% accuracy. As such, we also compute 2-class scores based on a threshold at the dataset mean, reported in the right-hand column of \ref{fig:2class}.

Keeping in mind the coarse nature of 2-class evaluation procedure, we wanted to use additional metrics to show the performance of each model. As introduced in previous work \cite{NIMA} LCC (linear correlation coefficient) and SRCC (Spearman rank correlation coefficient) are useful metrics for determining the correlation of a model's performance to the ground truth.

Earth movers distance is also an available metric for the task, and given it is the basis for the loss function we used for training, we also evaluate the models using pure EMD as a metric.

\subsection{Training}

Each model was trained using extensive hyperparameter sweeps and fine-tuning using Bayesian hyperparameter optimization techniques implemented in Weights and Biases. Often adding more hyperparameters to tune is viewed unfavorably since with grid and random sweeps this can increase time complexity of the sweeps significantly. This encumbrance was limited by Bayesian hyperparameter optimization leading to much faster hyperparameter convergence than with other sweeping methods. By using consumer-grade GPUs, we were limited to mini-batches of size 50, much smaller than those used in state-of-the-art methods. While NIMA never released their code or reported their hyperparameters, other re-implementations that would approach their accuracy used batch sizes of 96-100. With the complex problem of aesthetic assessment, batch size is an important factor in generalization when training a network. Due to this, our hardware limitations kept us from achieving their reported accuracy.   
During our sweeps we would tune the number of layers frozen with the use of Bayesian hyperparameter optimization, allowing the network to "choose" what amount of the network should be frozen with the pretrained weights, and what amount of the network should be allowed to freely train for the transfer task. Early on we found that the network preferred 85 - 95\% of the later layers in the network frozen, and as we proceeded with updating pretraining and checkpoints, the network would prefer to "thaw" more of the later layers for training to expand network capacity. 

Our final networks were first trained with SGD for 100 epochs until an EMD at or below 0.082 was achieved depending on the modality. After that, these initial checkpoints were used as pretraining using the same freezing methods and training runs of 10-20 epochs were started using the Adam optimizer \cite{Adam}. This step was repeated until no meaningful improvements were being made and the network was optimized as possible on the hardware available.

\subsection{Results and Analysis}

\begin{table}[h]
\begin{tabular}{lll} \toprule
    {Modality} & {\% Accuracy: 5} & {\% Accuracy:  Mean}  \\ \midrule
    Baseline  & 70.82 & 50.00 \\
    RGB  & 78.58 & 73.46 \\
    Depth  & 72.88  & 63.43   \\
    Blur  & 73.30  & 63.94 \\
    Saliency & 75.77  & 70.11 \\ \bottomrule
\end{tabular}
\caption{Table showing 2 class accuracy scores for each modality with a threshold of 5, and when the threshold is at the dataset mean. The baseline row represents what percent of the ground truth scores fall above the threshold.}
\label{fig:2class}
\end{table}

Table \ref{fig:2class} lists the two-class accuracy for each model, with thresholding at 5, and the dataset mean. With both these thresholds, saliency did best, followed by blur and depth. The ``Baseline" result shows that about 20\% of the distribution of the ground truth scores falls between 5 and the mean, and scores completed using a threshold of 5 significantly higher.

\begin{table}[H]
\begin{tabular}{lllll} \toprule
    {Modality} & {Mean} & {Mean } & {STD} & {STD }  \\ 
    {} & {LCC} & {SRCC} & {LCC} & {SRCC}  \\ \midrule
    RGB  & 0.652 & 0.643 & 0.641 & 0.6313 \\
    Depth  & 0.4099  & 0.3914 &  0.3969  & 0.3804  \\
    Blur  & 0.4165  & 0.4035 & 0.4072  & 0.3948 \\
    Saliency & 0.5651  & 0.5543 &  0.5545  & 0.543  \\ 
    NIMA & 0.636  & 0.612 & 0.233  & 0.218  \\  
    Pool-3FC & 0.756  & 0.757 & - & - \\ \bottomrule
\end{tabular}
\caption{Table showing LCC and SRCC correlation scores for each modality. Missing values were not provided from paper and were not able to be reproduced}
\label{fig:cctab}
\end{table}

Table \ref{fig:cctab} shows the LCC and SRCC reported from each modality and tells us a similar story to the accuracy evaluation. We can observe that the best correlation scores of the modalities was achieved by the saliency model, echoing the results shown in Table \ref{fig:2class}. We also report metrics achieved by NIMA \cite{NIMA} (the basis for our implementation) and Pool-3FC, the method achieving state of the art results on correlation metrics \cite{POOL3FC}. While we did not reach the mean correlation accuracy of Pool-3FC \cite{POOL3FC}, the additional loss terms allowed us to exceed NIMA's scores on the correlation metrics on RGB inputs.

\begin{table}[H]
\begin{tabular}{lll} \toprule
    {Modality} & {Validation EMD} & {Test EMD}  \\ \midrule
    RGB  & 0.06814 & 0.06828 \\
    Depth  & 0.08107  & 0.08130   \\
    Blur  & 0.08085  & 0.08050 \\
    Saliency & 0.07398  & 0.07377 \\ \bottomrule
\end{tabular}
\caption{Table showing base EMD loss achieved from fine tuning each model.}
\label{fig:emdtab}
\end{table}

In Table \ref{fig:emdtab} we report the EMD loss achieved on our validation and test sets.

\begin{table}[H]
\begin{tabular}{llll} \toprule
    {Loss Terms} & {Validation} & {2-Class} & {2-Class}  \\
    { }          & {EMD Loss}   & {(5.0)}  & {(5.38)}  \\ \midrule
    Pure EMD                                & 0.08461 &  72.1456 & 58.1829     \\
    EMD $ * \Delta \mu $                    & 0.08424          & 72.2795 & 57.6235  \\ 
    EMD $ * \Delta \sigma^2 $               & 0.08395     & 72.2638 & 58.5533  \\ 
    EMD $ * \Delta \mu * \Delta \sigma^2 $  & 0.08375 & 72.2480 & 57.9386 \\ \bottomrule
\end{tabular}
\caption{Final validation loss and 2 class accuracy scores for short training sweeps on models with and without loss terms.}
\label{fig:abltab1}
\end{table}
\begin{table}[H]
\begin{tabular}{lllll} \toprule
    {Loss Terms}  & {Mean} & {Mean} & {STD} & {STD} \\
    { }  & {LCC} & {SRCC} & {LCC} & {SRCC} \\ \midrule
    Pure EMD                                &  0.4994 & 0.4918 & 0.4934 & 0.4864     \\
    EMD $ * \Delta \mu $                    & 0.4715 & 0.4633 & 0.4659 & 0.4586 \\ 
    EMD $ * \Delta \sigma^2 $               & 0.4604 & 0.4544 & 0.4552 & 0.4492   \\ 
    EMD $ * \Delta \mu * \Delta \sigma^2 $  & 0.4309 & 0.4244 & 0.4275 & 0.4218   \\ \bottomrule
\end{tabular}
\caption{Final correlation metrics for short training sweeps on models with and without loss terms.}
\label{fig:abltab2}
\end{table}
% \begin{table}[htb]
% \begin{tabular}{Xlllll} \toprule
%     {Loss Terms} & {Validation} & {Mean} & {Mean} & {Var} & {Var} \\
%     { } & {EMD Loss} & {LCC} & {SRCC} & {LCC} & {SRCC} \\ \midrule
%     Pure EMD  & 0.08461 &  0.49944 & 0.4918 & 0.2270 & 0.2220     \\
%     EMD $ * \Delta \mu $ & 0.08424  & 0.4715 & 0.4633 & 0.2254 & 0.2218\\ 
%     EMD $ * \Delta \sigma^2 $ & 0.08395 & 0.4604 & 0.4544 & 0.2177 & 0.2145   \\ 
%     EMD $ * \Delta \mu * \Delta \sigma^2 $ & 0.08375 & 0.4309 & 0.4244 & 0.2111 & 0.2079   \\ \bottomrule
% \end{tabular}
% \caption{Final validation loss for short training sweeps on models with and without loss terms}
% \label{fig:res}
% \end{table}
% \\
% \begin{table}[htb]
% \begin{tabular}{XXXXX} \toprule
%     {Metric}                & {Pure EMD} & {EMD $ * \Delta \mu $} & {EMD $ * \Delta \sigma^2 $} & {EMD $ * \Delta \mu * \Delta \sigma^2 $} \\ \midrule
%     Val EMD Loss     & 0.08461 &  0.08424 & 0.08395 & 0.08375     \\
%     2-Class(5)      & 72.1456 & 72.2795 & 72.2638 &  72.24804 \\
%     2-Class(5.38)   & 58.1829 & 57.6235 & 58.5533 & 57.9386 \\
%     Mean LCC                & 0.49944  & 0.4715 & 0.4604 & 0.4309 \\
%     Mean SRCC               & 0.4918 & 0.4633 & 0.4544 & 0.4244    \\ 
%     Var LCC                 & 0.2270 & 0.2254 & 0.2177 & 0.2111    \\
%     Var SRCC                & 0.2220 & 0.2218 & 0.2145 & 0.2079 \\ \bottomrule
% \end{tabular}
% \caption{Final validation loss for short training sweeps on models with and without loss terms}
% \label{fig:res}
% \end{table}

To study the effect of the loss terms we set up a limited sweep with each just EMD, EMD with either the mean or variance terms, and with the full combined loss function. Each was provided the same pretraining for 100 epochs on the standard AVA dataset. The sweep was then started by locking hyperparameters to the most successful ones used in fine tuning, however the learning rate for the convolution and dense layers as well as the learning rate decay rate was left with some room to adjust. Each run was allowed to run 10 epochs.  The results shown are after each sweep completed 12 runs, with the lowest final validation loss values out of all of the runs shown in the table.  When we are viewing the results in Table \ref{fig:abltab1}, our loss terms provide a gradual improvement to the evaluation scores. Viewing the 2 class accuracy scores, we can see that just the variance term being used yielded the best score, with the combination coming in second. Finally the correlation scores in Table \ref{fig:abltab2} show that using a pure EMD loss is best, and the additional loss terms decrease the correlation. While the models' training has not been fully completed, these results show us that there is a complex relationship and tradeoff between EMD loss, classification accuracy, and correlation.

Below we show example groups of images from the validation set that models preferred, meaning that out of the three modalities, the model trained for the specific modality achieved an earth movers distance the closest to the ground truth. These groupings represent an example set of the possible automatic labeling that can be completed using our method, and display differences in the images that the modalities prefer. 
\\

\begin{figure}[H]

\centerline{\includegraphics[scale=0.17]{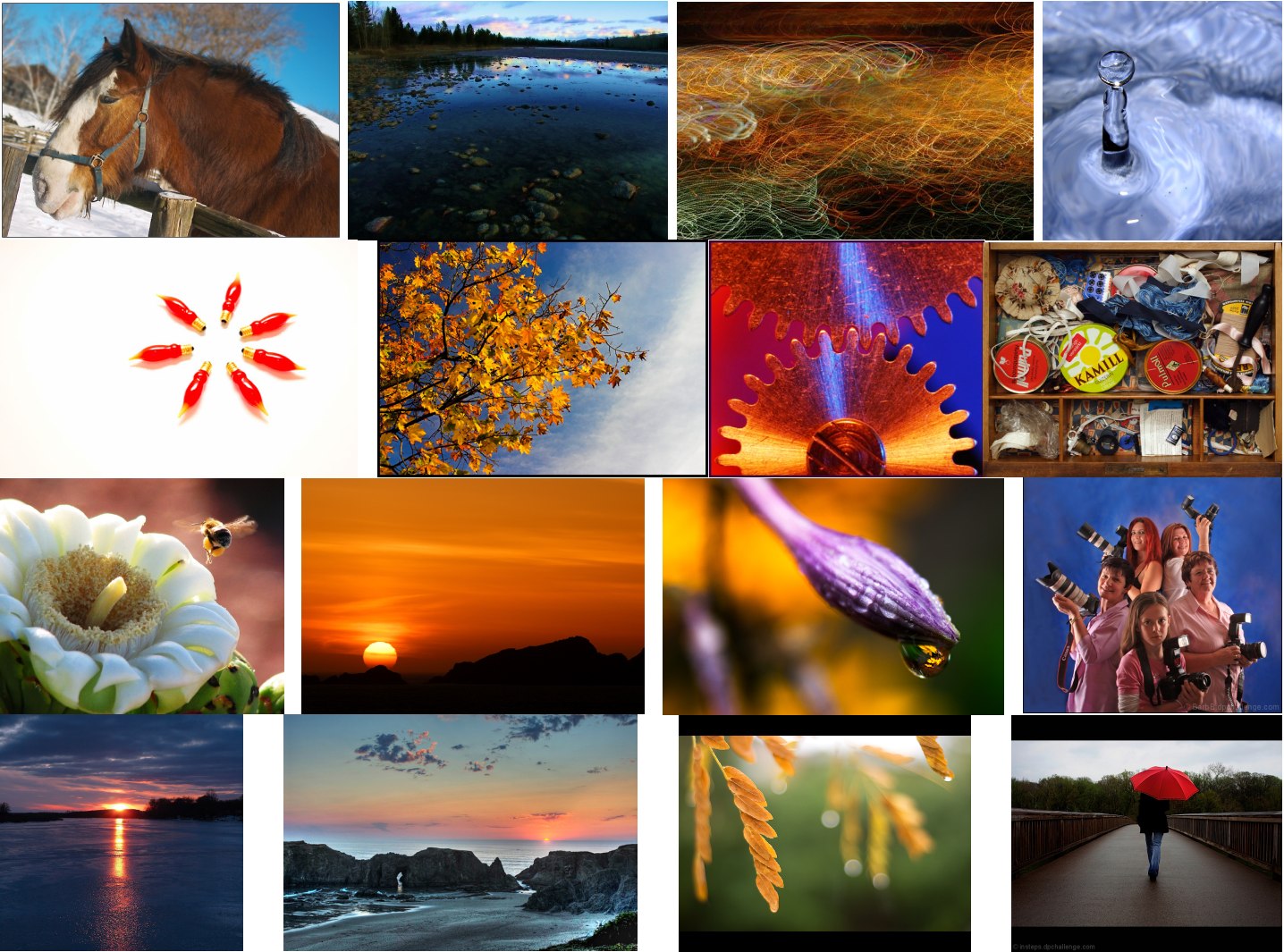}}
\caption{Blur Examples}
\label{fig:blurex}
\end{figure}

\begin{figure}[H]
\centerline{\includegraphics[scale=0.17]{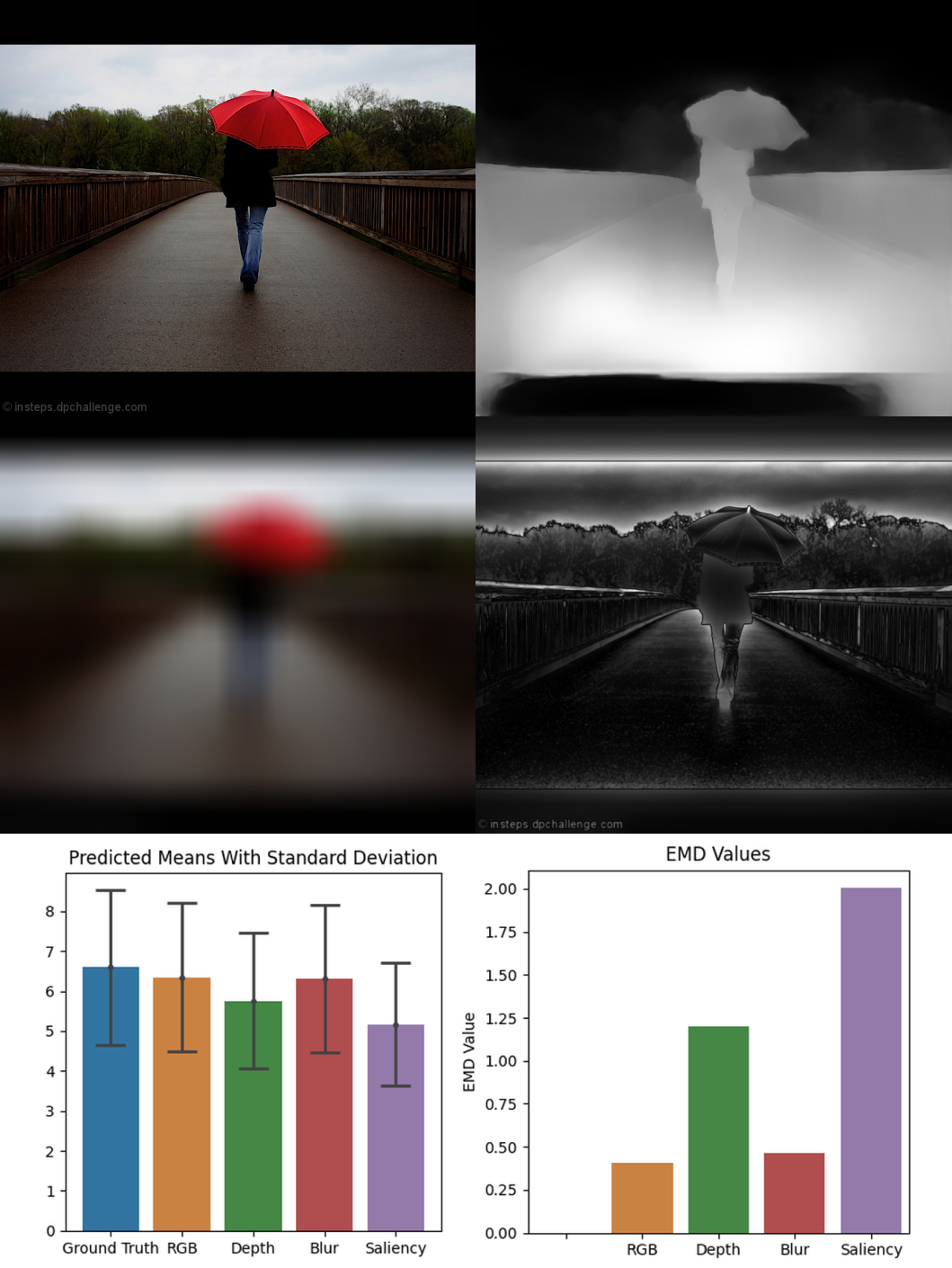}}
\caption{Blur Details}
\label{fig:blurdet}
\end{figure}

\begin{figure}[H]

\centerline{\includegraphics[scale=0.17]{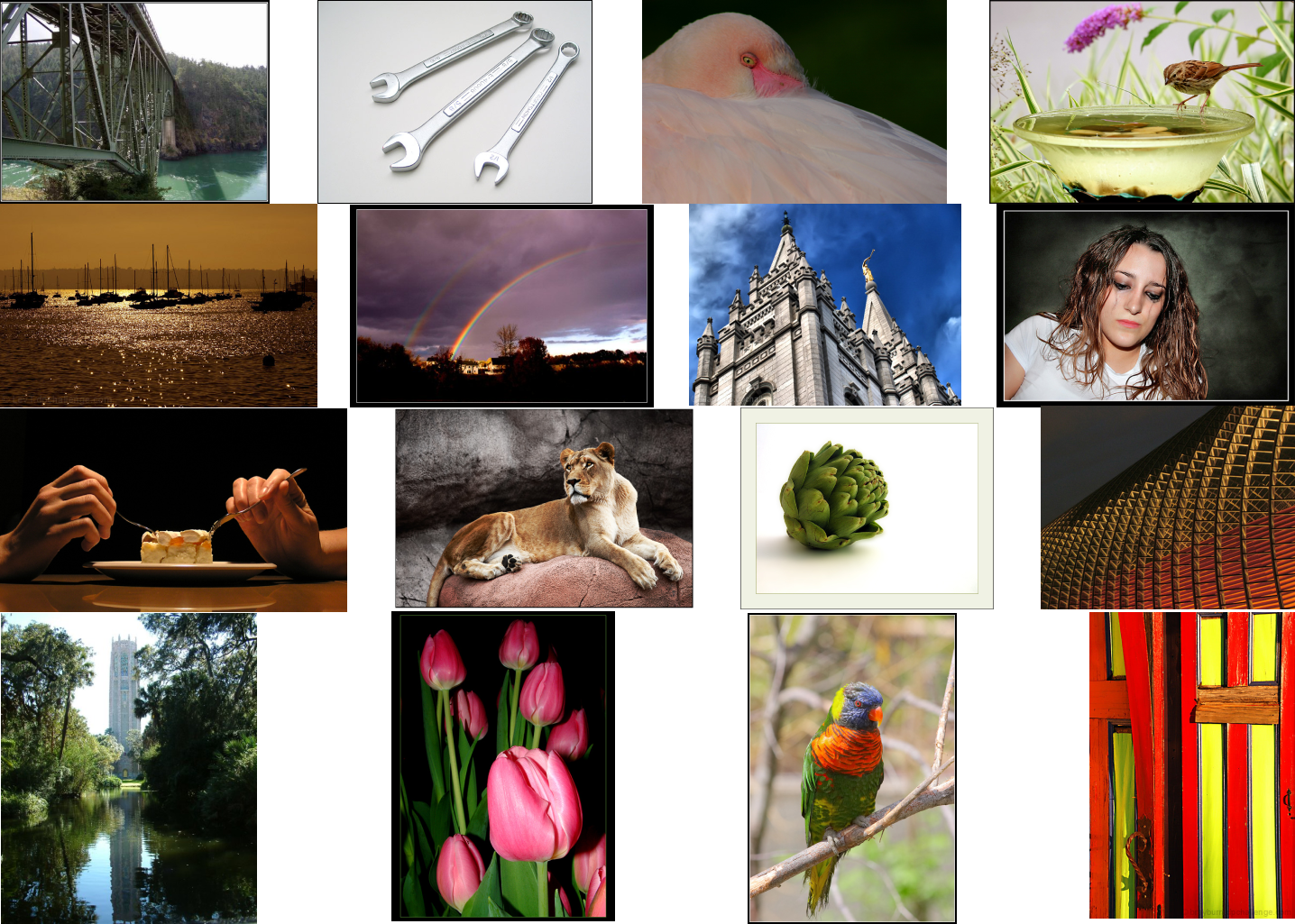}}
\caption{Depth Examples}
\label{fig:depthdex}
\end{figure}

\begin{figure}[H]
\centerline{\includegraphics[scale=0.18]{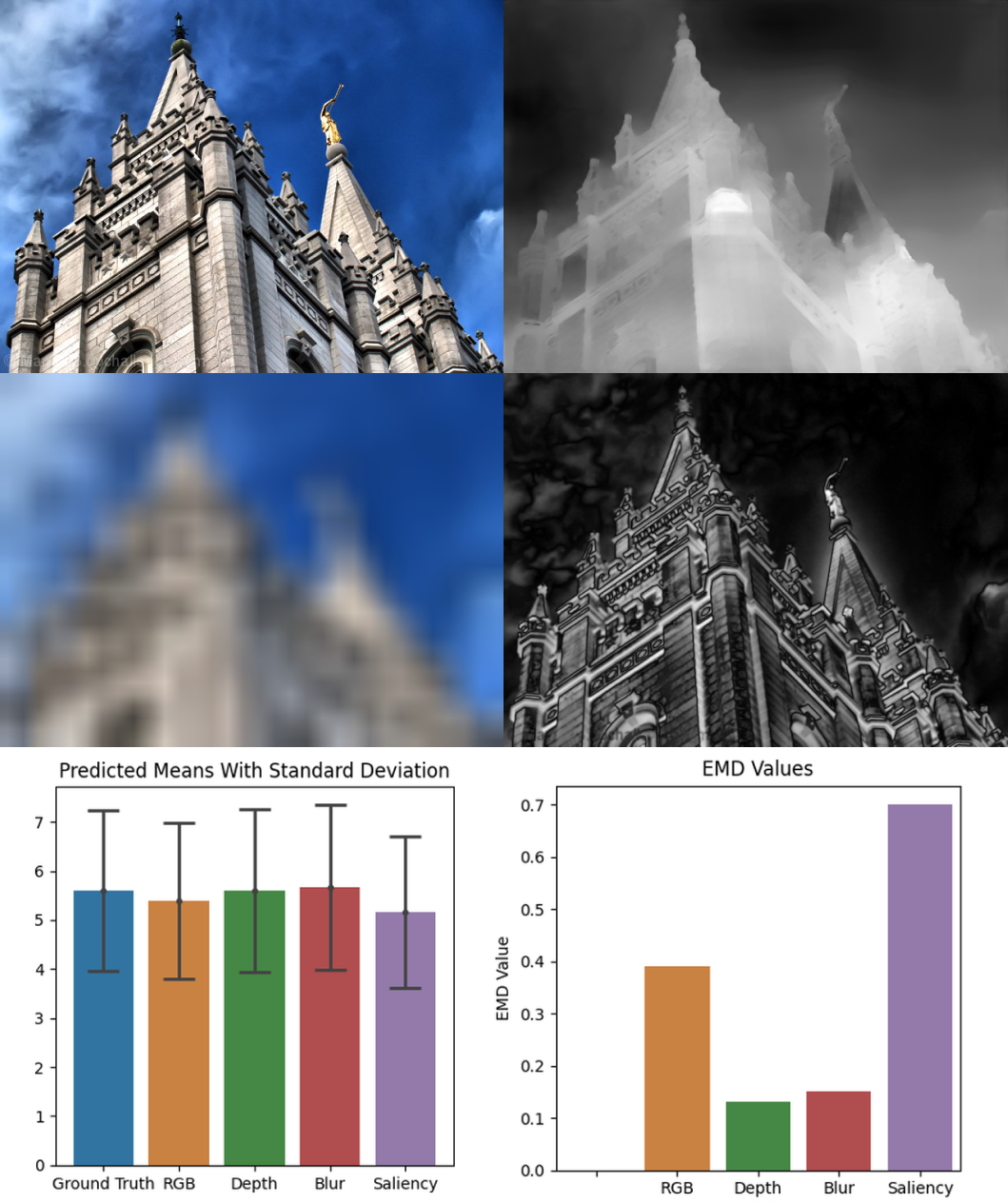}}
\caption{Depth Details}
\label{fig:depthdet}
\end{figure}

\begin{figure}[H]
\centerline{\includegraphics[scale=0.17]{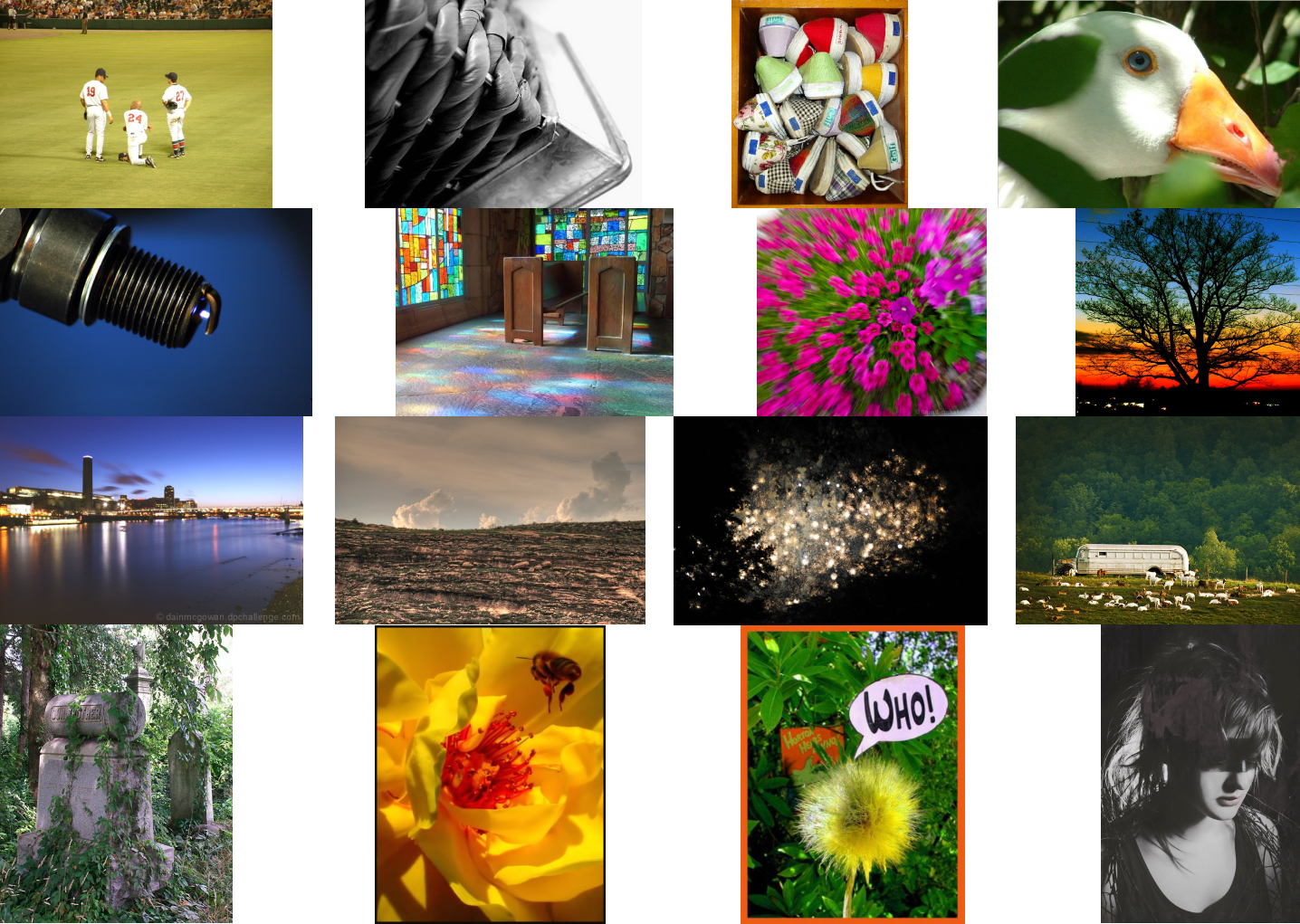}}
\caption{Saliency Examples}
\label{fig:salex}
\end{figure}

\begin{figure}[H]
\centerline{\includegraphics[scale=0.17]{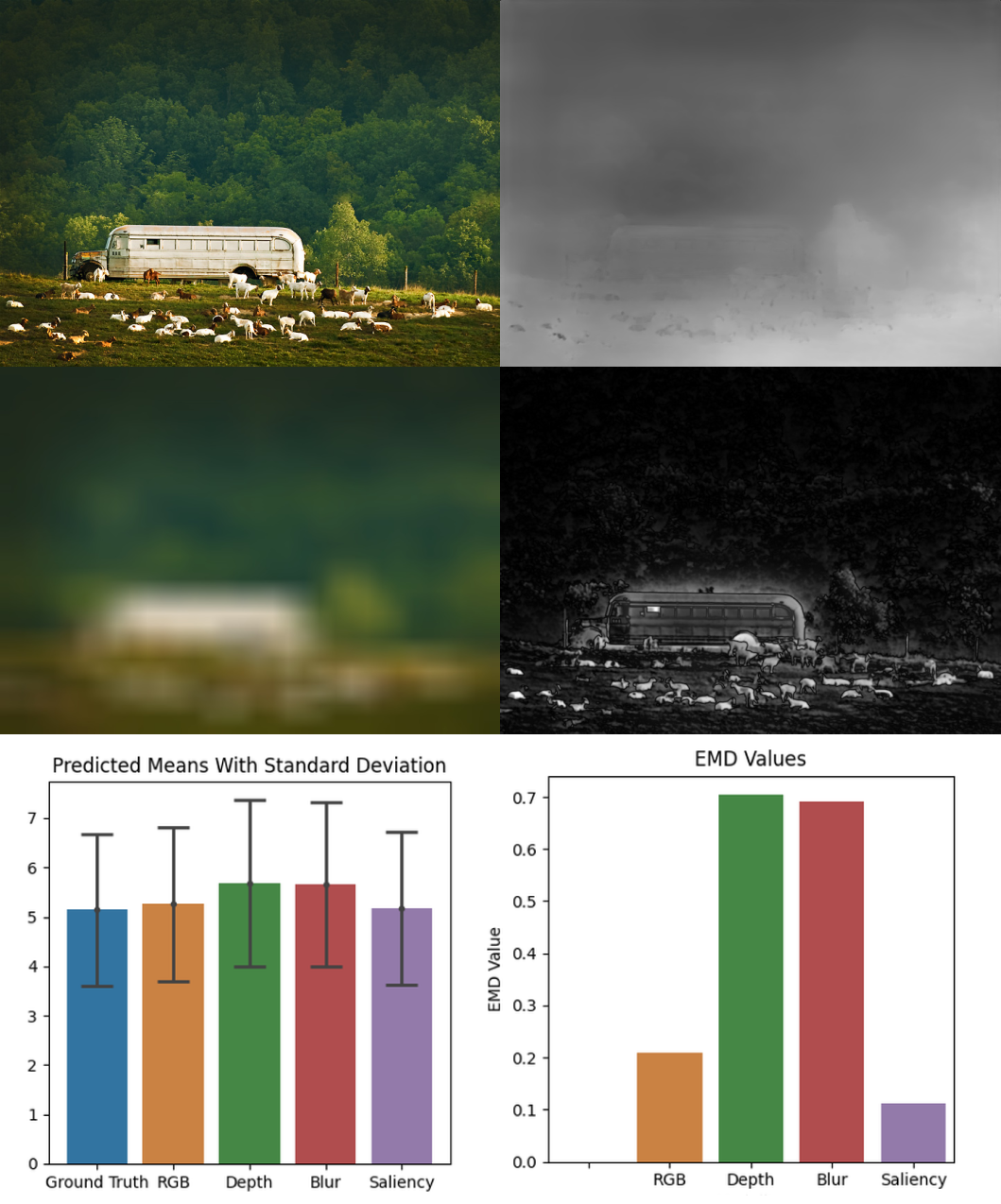}}
\caption{Saliency Details}
\label{fig:saldet}
\end{figure}

\begin{figure}[H]
\centerline{\includegraphics[scale=.5]{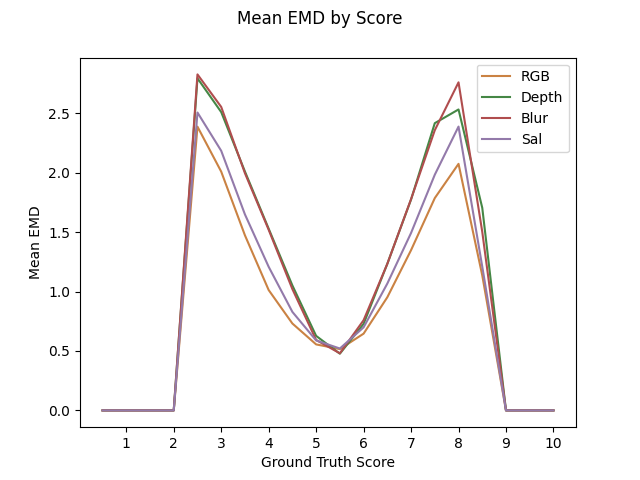}}
\caption{We group EMD scores based on their ground truth aesthetic score for that datapoint and plot the mean}
\label{fig:scorebins}
\end{figure}

In order to further examine the results of each network, we wanted to visualize the relationships between the ground truth mean score for each image, and how that impacted the quality of predictions. In Figure \ref{fig:scorebins}  we can see this visualization for each modality. Each prediction in the validation set was put into bins according to its ground truth score with its corresponding EMD recorded. In this figure, we can observe that the saliency modality does significantly better then the other two as well as the significant dip around the score of 5. This dip further displays aesthetic model's dispositions towards predicting the dataset mean, which is likely a direct impact of the ground truth scores of the images also being grouped around 5. Other than that, there is not much difference between the modalities on this macro scale. This lack of difference in the modalities, reflected in other analysis begins to show us that the different modalities are picking up on common or similar latent aesthetic content. 

\begin{figure}[H]
\includegraphics[scale=.25]{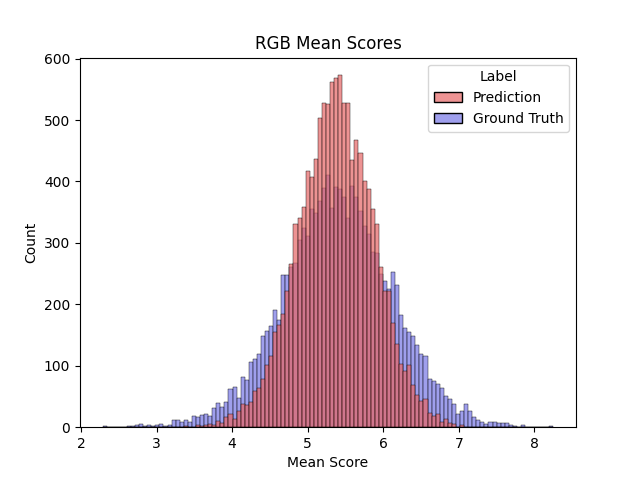}
\includegraphics[scale=.25]{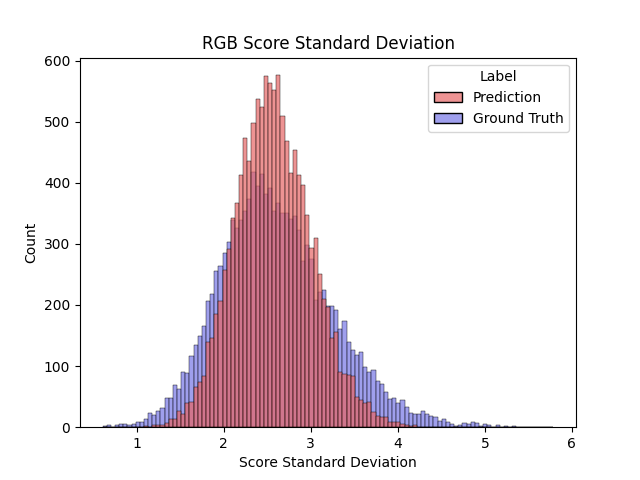}

\includegraphics[scale=.25]{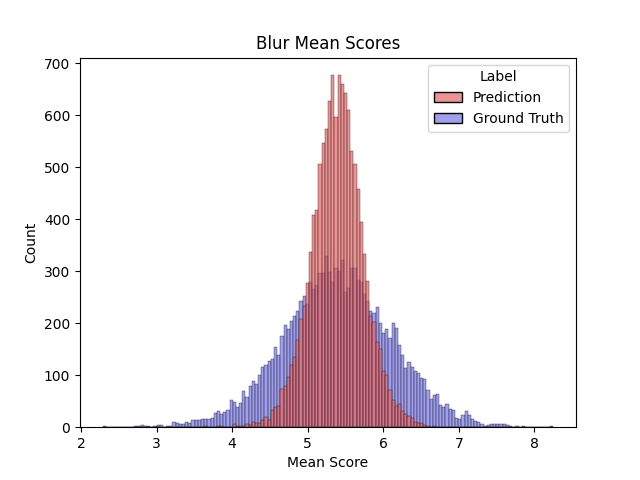}
\includegraphics[scale=.25]{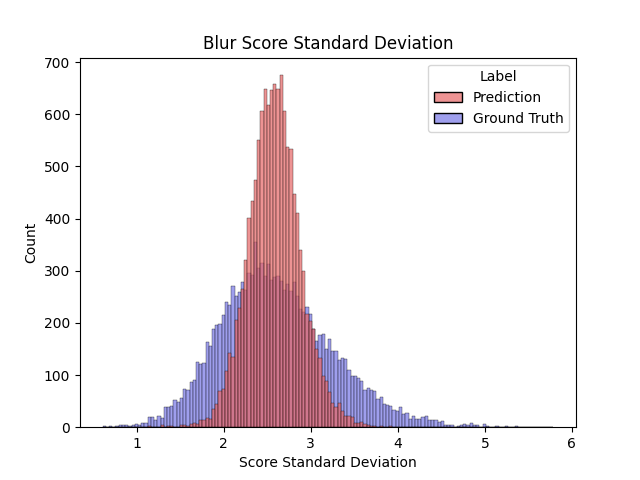}

\includegraphics[scale=.25]{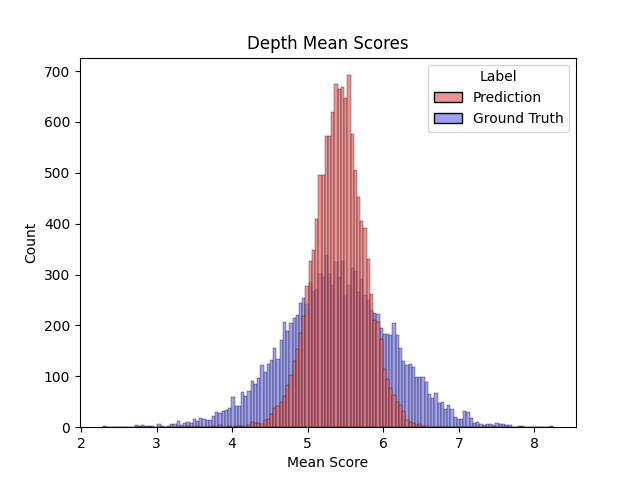}
\includegraphics[scale=.25]{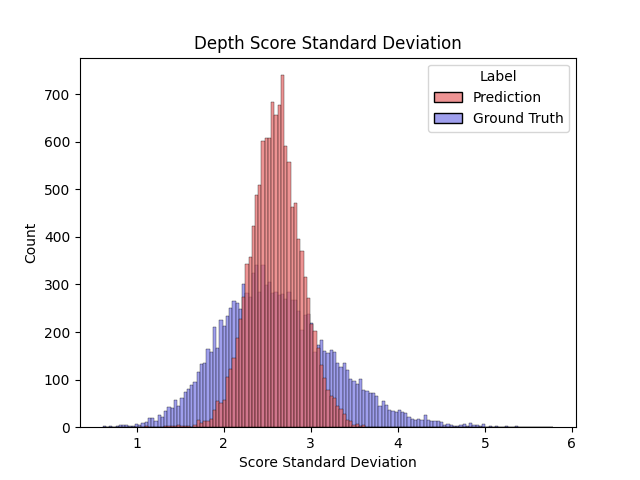}

\includegraphics[scale=.25]{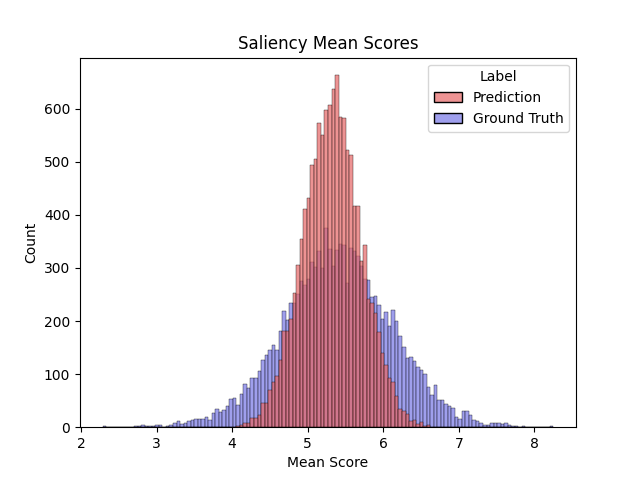}
\includegraphics[scale=.25]{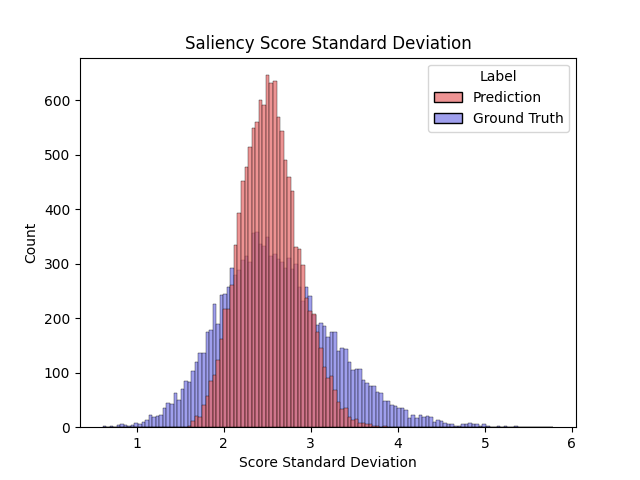}
\caption{Mean and Standard Deviations of predictions plotted against ground truth for each modality }
\label{fig:hist}
\end{figure}

\begin{table}[H]
\begin{tabular}{lll} \toprule
    {Modality} & {$\mu$ Histogram EMD} & {$\sigma ^2$ Histogram EMD}  \\ \midrule
    RGB  & 414.1600 & 451.8286 \\
    Depth  & 777.3568  & 853.7433   \\
    Blur  & 770.3130  & 837.7480 \\
    Saliency & 510.9968 & 559.1916 \\ \bottomrule
\end{tabular}
\caption{Table showing EMD distance between predicted and ground truth means and standard deviations for each modality.}
\label{fig:histtab}
\end{table}

In Figure \ref{fig:hist}, we can see a visual display of each modalities' predicted means as well as their standard deviations and how they compare to the ground truth distributions. As seen in previous work \cite{NIMA}, the mean scores from the original AVA dataset match the ground truth distribution quite well. The problem with previous work is the standard deviation distributions did not fit well: with our method we see a better balance in achieving both tasks. We can also observe that the baseline RGB and saliency models while having the highest metric scores also visually "fit" the ground truth scores better than the other modalities. This is reinforced by Table \ref{fig:histtab} where EMD scores were calculated between the two distributions as displayed in Figure \ref{fig:hist}. We can see in these modalities that do not fit as well, that the predictions are tightly clustered around the dataset mean, further guiding the conclusion that a carefully chosen threshold is important for the two class accuracy metric when assessing aesthetic models.

\begin{table}[hbtp]
\begin{tabular}{lll} \toprule
    {Modality} & {\% Accuracy: 5} & {\% Accuracy:  Mean}  \\ \midrule
    RGB  & 62.556 & 57.259 \\
    Depth  & 52.300  & 42.501   \\
    Blur  & 52.058  & 51.477 \\
    Saliency & 55.000  & 61.4 \\ \bottomrule
\end{tabular}
\caption{Table showing 2 class accuracy scores for each modality with a threshold of 5, and when the threshold is at the dataset mean for the AADB dataset.}
\label{fig:aadbtab}
\end{table}

We wanted to evaluate our methods trained on AVA on a completely separate dataset, in this case AADB \cite{AADB} a dataset of images from Flickr represents images that are collected and labeled differently from AVA at a different time. From Table \ref{fig:aadbtab} we can see the accuracies are lower than our validation and testing set, but they still show similar results relative to each other even on out of distribution data. We gather from the results on an-out of distribution dataset suggest that aesthetic scoring is strongly linked to the year the dataset was collected or the audience scoring it, suggesting common aesthetic features between datasets can be sparse. 

\subsection{Covariance Analysis}

In order to assess the relationships between our modalities and other hand-designed categories, we created a series of covariance visualizations that show how performance on each of our modalities relate to each of a given set of category labels or ratings. 

For these covariance calculations we compare the EMD of each prediction in a category to either the score given in that category, or a one-hot vector to represent a classification. Since with EMD a lower distance is better, we negated these scores so that higher is better. This provides us with a visualization where a higher correlation score means the model performed well relative to that category. 

First, we compared our model's EMD outputs to the content and aesthetic category labels available in the subset of data points re-labeled by EVA \cite{EVA}. Since our modalities do not align directly with EVA's categories, we do not expect a direct correspondence between any of our modalities and the categories proposed in EVA. However, we can spot some trends in which modalities relate to which categories based on their covariances. The EVA categories are defined as follows:

\begin{itemize}
\item Light and color ("visual") relates to visual perception, including brightness, contrast, and color saturation. 
\item Composition and depth relates to the position and spatial relationship between objects in the scene.
\item Quality can be impacted by different types of distortions, including blur, compression, noise, and other artifacts. 
\item Semantics is related to how much the subject subjectively likes the content of the image. 
\end{itemize}

Each aesthetic category in EVA is labeled with a score of 1-4. 

\begin{figure}[hbtp]
\centerline{\includegraphics[scale=.5]{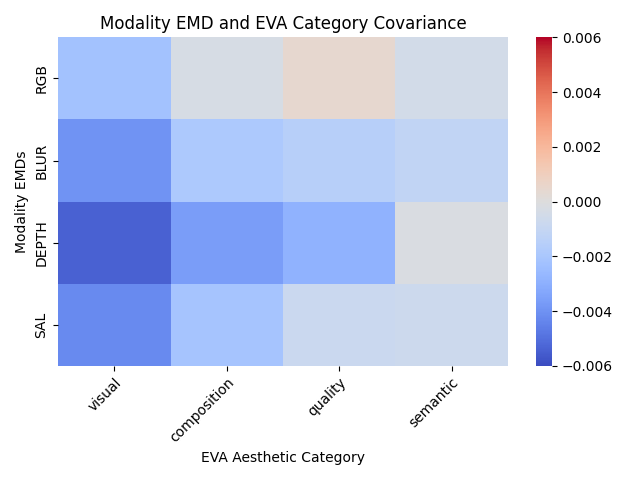}}
\caption{Covariance matrix showing the relation of our modality predictions to EVA aesthetic category labels. Positive correlation indicates that a modality did a good job in predicting a category's score.}
\label{fig:evaath}
\end{figure}

\begin{figure}[hbtp]
\centerline{\includegraphics[scale=.5]{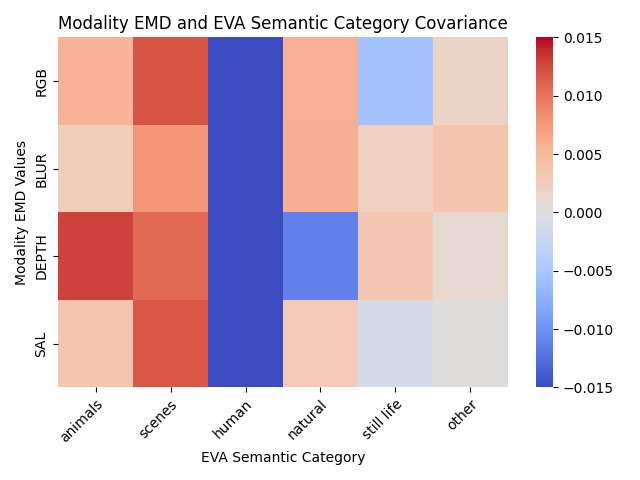}}
\caption{Covariance matrix showing the relation of our modality predictions to EVA content category labels. Positive correlation indicates that a modality did well in predicting a category's score.}
\label{fig:evasub}
\end{figure}

In Figure \ref{fig:evaath}, we can see that the red colors represent values around zero correlation, however we see some drift into negative correlations when viewing the depth model performance, as well as images with the "visual" label as described above. With depth preserving the least amount of lighting and color information this negative correlation matches assumptions that can be made about the relationship between the two. We see the highest correlation between EVA's quality category and our original RGB model, with the second highest being the saliency model. This tracks as these are the two models that preserve or enhance the fine grained details in the images. Another point lacking negative correlation is between the depth modality and semantic category, since depth often highlights the shape and form of objects in the foreground, this correlation is also significant. While the correlation scores are relatively small, they still give us insight into the different aspects each modality highlights throughout the dataset.
\\

In Figure \ref{fig:evasub}, we observe relatively high correlation scores between the depth modality and animal subjects, and between scenes and the saliency and RGB modalities. Also a low correlation in most of the images with human subjects is present, with the lowest from the blur modality. Given possible bias and sensitive nature labeling of human subjects can be, we can assume that aesthetic models do not perform well when presented with a range of human subject matter. From this visualization we can begin to observe patterns in preference between the models similar to observations in Figures \ref{fig:blurex} - \ref{fig:saldet}.
\begin{figure}[hbtp]
\centerline{\includegraphics[scale=.5]{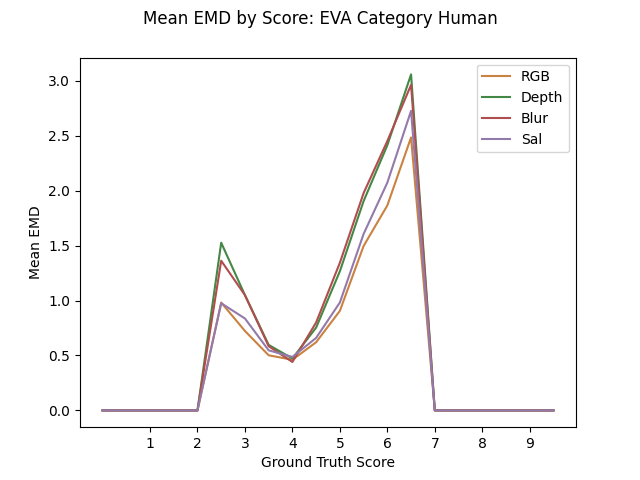}}
\centerline{\includegraphics[scale=.5]{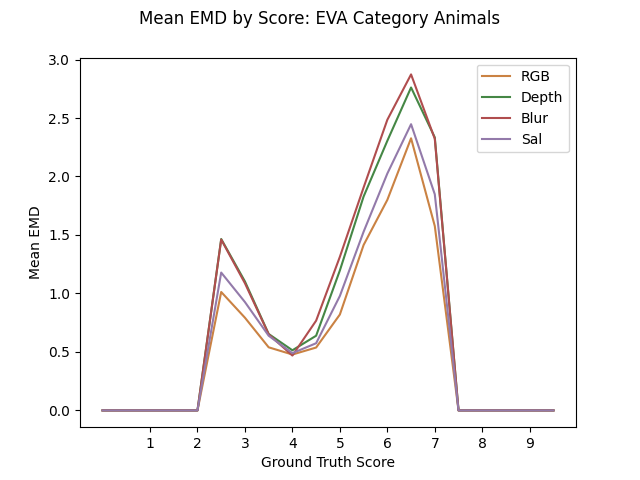}}
\centerline{\includegraphics[scale=.5]{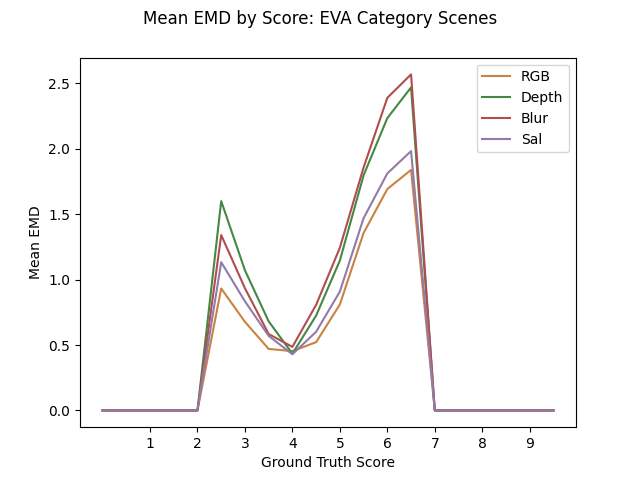}}
\caption{Mean EMD as a function of ground truth score, however each graph represents one of the EVA semantic subsets highlighted by Figure \ref{fig:evasub}.}
\label{fig:evascorebins}
\end{figure}

To validate and further analyze these relationships seen in the EVA categories we generated EMD values as a function of ground truth score again, except specific to the EVA categories of note as seen in Figure \ref{fig:evasub}. We see the relationships mirrored from the covariance matrix in the relative performance of the modalities in Figure \ref{fig:evascorebins}. In Figure \ref{fig:evasub} we observed that all modalities received negative correlation scores for human subjects, and in our breakdown by score for the human category the gap between each modality is smaller than in the others, as well as depth and blur both did poorly on both the highest and lowest of scores. In both the animal and scene specific breakdowns we can observe the depth modality performing better than blur as is mirrored in the positive correlation scores. From these results we can conclude that the covariance analysis does an adequate job at assessing the performance of each model in these categories, with the added information of seeing the most deviation in the highest and lowest scoring images.

% \begin{table}[htb]
% \begin{tabular}{Xllll} \toprule
%     {Metric} & {RGB} & {Depth} & {Blur} & {Saliency}  \\ \midrule
%     \% Corrent From 5  & 62.556 & 52.300 & 52.058 & 53.400  \\
%     \% Correct From Mean  & 57.259  & 57.499 & 51.477  & 51.7   \\ \bottomrule
% \end{tabular}
% \caption{Table showing 2 class accuracy scores for each modality with a threshold of 5, and when the threshold is at the dataset mean for the AADB dataset.}
% \label{fig:res}
% \end{table}

\begin{figure}[hbtp]
\centerline{\includegraphics[scale=.5]{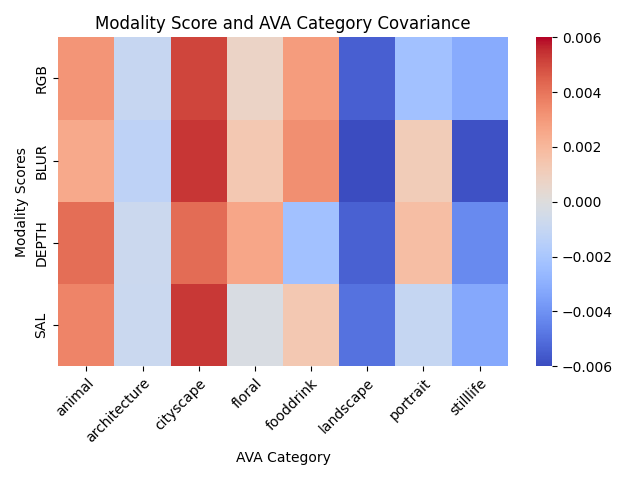}}
\caption{Covariance matrix showing the relationship between our modality prediction and original AVA categories. Positive correlation indicates that a modality did well in predicting a category's score.}
\label{fig:avacat}
\end{figure}

\begin{figure}[hbtp]
\centerline{\includegraphics[scale=.5]{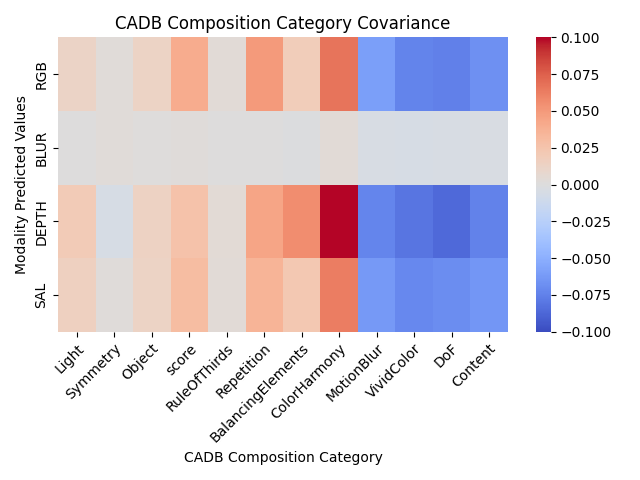}}
\caption{Covariance matrix showing the relationship between our modality predictions from the CADB dataset and corresponding composition categories. Positive correlation indicates that a modality did well in predicting a category's score. Distances were measured between ground truth and predicted mean.}
\label{fig:cadbcat}
\end{figure}

Next we completed a similar covariance analysis on AVA and its original categories provided by the photography challenges where the dataset was originally collected from. This can be seen in Figure \ref{fig:avacat}. In this figure we can see that each of the modalities did well in cityscape and animal categories, and worse in images containing landscape and still life scenes. Looking at the blur modalities' row, we can observe some of the highest and lowest scores, suggesting that the blur was the most polarizing transformation when concerning the cityscape, landscape, and still life subjects. When we look at the floral and portrait columns we see depth having the highest correlation scores, this can be interpreted again as depth focusing on images with strong foreground shapes and structure, both important factors in these types of subjects.  

Finally we completed a similar experiment with the CADB dataset \cite{CADB}. The CADB dataset however uses a different range of scores and therefore different number of bins for its distributions. For this dataset we compared the difference between predicted and ground truth means, and multiplied by negative one, so that positive correlations indicate the model did well for that category as seen in Figure \ref{fig:cadbcat}. From the figure the first obvious thing is the blur row's correlation is consistently close to zero. This suggests that the blur modality is successful in being composition agnostic, leaving little compositional information behind after the blurring process is complete. We notice similar vertical zero-correlation strips for the rule of thirds and symmetry categories, however this time their orientation suggests that these factors are either not well represented by any of our modalities, or were not significant aesthetic features picked up by our models or lacking presence in the AVA dataset. 

When viewing these covariance results, some of these correlations make sense, e.g. in Figure \ref{fig:evaath} seeing "visual" (the category relating to color and light) be negatively correlated with a greyscale, depth modality. At the same time others do not, like the highest positive correlation in Figure \ref{fig:cadbcat} where "Color Harmony" is showing positive correlation with the depth modality. It is important to remember when interpreting these results, that many factors are at play, often dealing with how well the image's content is captured by the modality, the accuracy of that model, as well as data points that are not evenly distributed in relation to the dataset mean or standard deviation for that specific category. While these results are surprising, they can also open further investigation into the significance of these hand-designed categories and how relevant they are to image aesthetics and composition. 

\section{DISCUSSION}

From our experiments and results we are able to make some conclusions about automatic aesthetic assessment and the modality breakdown methods we applied. It was surprising to find such significant similarities between the different modalities in preference. As we observed in both the EMD by score and the covariance analysis, the differences seen in the modality results are often small, and suggests that similar content and features were being discovered by the networks. The task of automatic image assessment is also complex and can be impacted by bias during the labeling process and dissenting opinion of subjective material. 

As we saw from our results, the saliency modality outperformed the other modalities in all metrics applied. Given that the generated images are greyscale, it raises questions about how much specific colors impact aesthetic assessment. Color harmony and color balance are established features in the aesthetic assessment space, and it is interesting to see how much can be done even without color. Visual saliency has been used in related work \cite{A-LAMP} to provide guidance for patch selection and has history in the field for algorithm-based automatic cropping techniques \cite{oldphotocomp}. 

In generating modalities it became clear that some of our methods would be limited by previous work done in generation of the saliency and depth modalities. These methods are established estimations of information we are interested in, and in a perfect world we would have ground truth depth, and hand-labeled visual saliency regions. Involving the blur modality as well, all modalities cannot capture what we are looking for in every image. The depth modality struggled with images with post processing applied, or that contained difficult content for depth estimation. Multiple modalities also struggled when images contained borders and watermarks.

\section{FUTURE WORK}

In the future we hope to apply our method to larger and more recent models supported by a corresponding increase in access to hardware. This would allow us to explore in greater depth the differences in modalities if they are present. If these models also fail to show a large macro difference between modalities, it would solidify our experiments' results that each modality shares much aesthetic information in common. 

Other work surrounds expanding and improving the modalities and even creating unique models for each modality. For example instead of blur we could decompose the image to a color histogram to completely throw out spatial information. Another example would be improved depth processing that works for every image, or an increased number of image examples.

AVA was collected and labeled over a decade ago. Over this time period the general public's perception and opinion of what makes a photo aesthetically good or bad has changed due to many factors. An updated large dataset with labels similar to the ones seen in EVA, AADB, and CADB would be of great use to the field.

\section{CONCLUSION}

Our method provides a unique and novel insight into a dominant and widely accepted aesthetics dataset. We show improvements in correlation scores over NIMA using hyperparameter controlled loss terms, and apply this method to generated modalities of the source dataset. We discussed weaknesses and improvements for common metrics. Our method provided us with insights into the relationships that the modalities highlighted as well as producing an automatic method with comparable results to related work for aesthetic categorization and explainability. Through this paper we hope to establish and groundwork and encourage future aesthetic papers to provide explainable results and transparent metrics as is critical to maintaining the integrity of aesthetic assessment.

\bibliographystyle{IEEEbib}
\bibliography{strings,refs}

\end{document}